\documentclass[%
 reprint,
superscriptaddress,
 amsmath,amssymb,
pre,
]{revtex4-1}

\usepackage{graphicx}
\usepackage{dcolumn}
\usepackage{bm}
\usepackage{lipsum}

\usepackage[caption=false]{subfig}

\begin{document}


\title{Mastering high-dimensional dynamics with Hamiltonian neural networks}

\author{Scott T. Miller}
\affiliation{Nonlinear Artificial Intelligence Laboratory, Physics Department, North Carolina State University, Raleigh, NC 27607, USA }

\author{John F. Lindner}
\email[Corresponding author: ]{jlindner@wooster.edu}
\affiliation{Nonlinear Artificial Intelligence Laboratory, Physics Department, North Carolina State University, Raleigh, NC 27607, USA }
\affiliation{Physics Department, The College of Wooster, Wooster, OH 44691, USA}

\author{Anshul Choudhary}
\affiliation{Nonlinear Artificial Intelligence Laboratory, Physics Department, North Carolina State University, Raleigh, NC 27607, USA }

\author{Sudeshna Sinha}
\affiliation{Nonlinear Artificial Intelligence Laboratory, Physics Department, North Carolina State University, Raleigh, NC 27607, USA }
\affiliation{Indian Institute of Science Education and Research Mohali, Knowledge City, SAS Nagar, Sector 81, Manauli PO 140 306, Punjab, India}

\author{William L. Ditto}
\affiliation{Nonlinear Artificial Intelligence Laboratory, Physics Department, North Carolina State University, Raleigh, NC 27607, USA }

\date{\today}

\begin{abstract}
We detail how incorporating physics into neural network design can significantly improve the learning and forecasting of dynamical systems, even nonlinear systems of many dimensions. A map building perspective elucidates the superiority of Hamiltonian neural networks over conventional neural networks. The results clarify the critical relation between data, dimension, and neural network learning performance. 
\end{abstract}

\maketitle 

\section{Introduction}
Artificial neural networks are powerful tools being developed and deployed for a wide range of uses, especially for classification and regression problems~\cite{Haykin2008}. They can approximate continuous functions~\cite{Cybenko1989, Hornik1991}, model dynamical systems~\cite{Lusch2018,Jaeger78,Pathak2018,Carroll2019}, elucidate fundamental physics~\cite{SciNet,AIFeynman,AIPhysicist}, and master strategy games like chess and Go~\cite{AlphaZero}. Recently their scope was extended by exploiting the symplectic structure of Hamiltonian phase space~\cite{HNN,HGN,mattheakis2019,Bertalan2019,Bondesan2019} to forecast the dynamics of conservative systems that mix order and chaos~\cite{choudhary}. 

Although \textit{recurrent} neural networks~\cite{Jaeger78,Pathak2018,Carroll2019} have been used to forecast dynamics, we study the more popular \textit{feed-forward} neural networks as they learn dynamical systems of increasingly high dimensions. The ability of neural networks to accurately and efficiently learn higher dimensional dynamical systems is an important challenge for deep learning, as real-world systems are necessarily multi-component and thus typically high-dimensional. Conventional neural networks often perform significantly worse when they encounter high-dimensional systems, and this renders them of limited use in complex real-world scenarios comprised of many degrees of freedom. So it is crucial to find methods that scale and continue to efficiently and accurately learn and forecast dynamics under increasing dimensionality.

In this work, we demonstrate the scope of Hamiltonian neural networks' ability to efficiently and accurately learn high-dimensional dynamics. We also provide an alternate \textit{map building} perspective to understand and elucidate how Hamiltonian neural networks (HNNs) learn differently from conventional neural networks (NNs). In particular we demonstrate the significant advantages offered by HNNs in learning and forecasting higher-dimensional systems, including linear and nonlinear oscillators and a coupled bistable chain. The pivotal concept is that HNNs learn the single energy surface, while NNs learn the tangent space (where the derivatives are), which is more difficult for the same training parameters. As the number of derivatives grow with the dimension, so too does the HNN advantage.

\begin{figure}[t!]
	\includegraphics[width=0.95\linewidth]{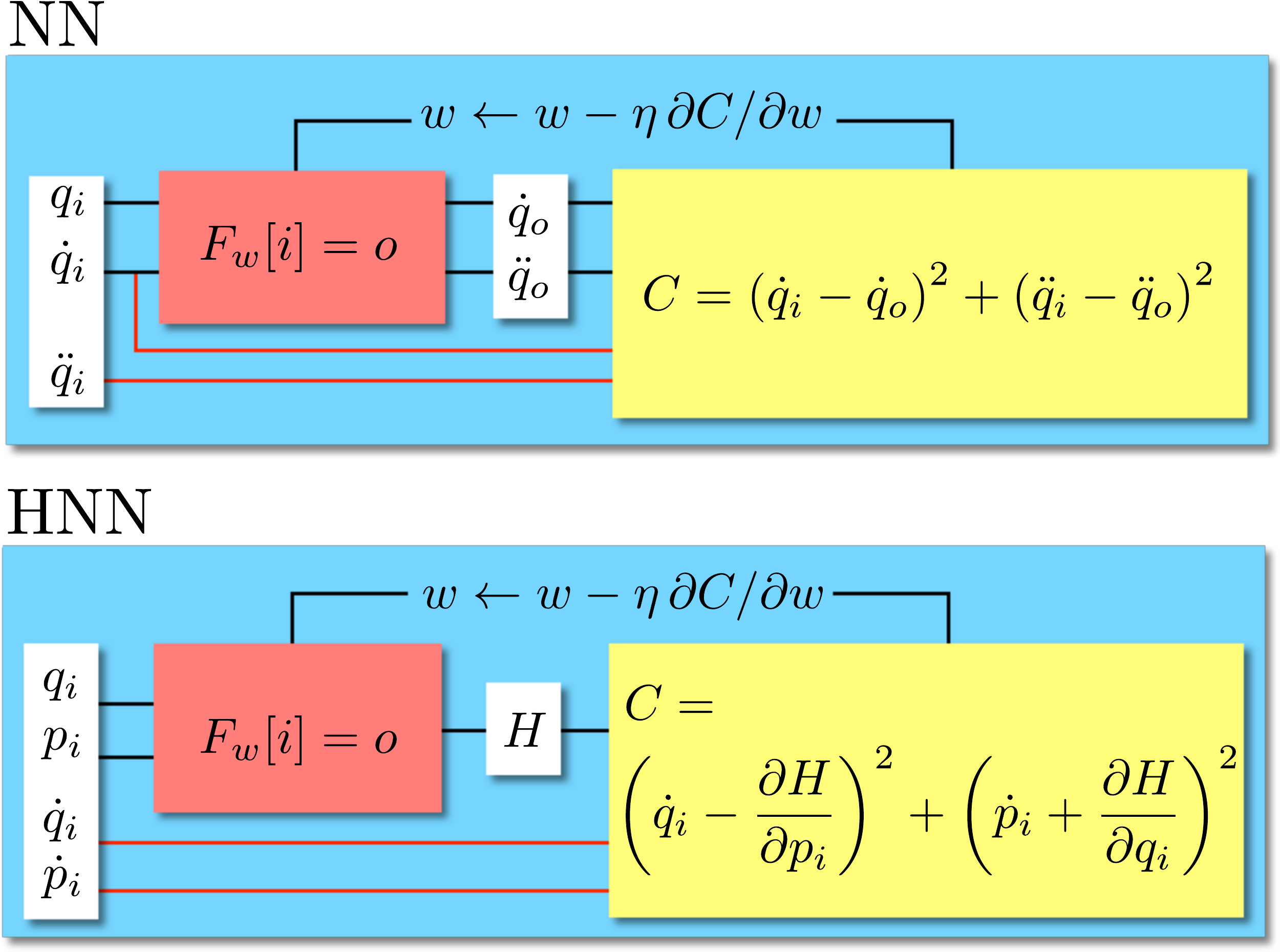} 
	\caption{Dynamics neural network schematics. Output $o$ of each network is a nonlinear function $F$ of its input $i$ and its weights and biases $w$, which adjust during training with learning rate $\eta$ to minimize a cost function $C$. Conventional neural network (NN) intakes positions $q_i$ and velocities $\dot q_i$, outputs velocities $\dot q_o$ and accelerations $\ddot q_o$, and needs input accelerations $\ddot q_i$ to compute costs (top). Hamiltonian neural network (HNN) intakes positions $q_i$ and momenta $p_i$, outputs only the energy $H$, but needs velocities $\dot q_i$ and forces $\dot p_i$ to compute costs (bottom).}
	\label{xNN}
\end{figure}

\section{Neural Networks}

A neural network is a nonlinear function
\begin{equation} \label{nonlinearEq}
   o=F[i,w]=F_w[i]
\end{equation}
that converts an input $i$ to an output $o$ according to its (typically very many) weights and biases $w$. Training a neural network with input-output pairs repeatedly updates the weights and biases by
\begin{equation}
    w \leftarrow w - \eta \frac{\partial C}{\partial w}
\end{equation}
to minimize a cost function $C$, where $\eta$ is the learning rate, with the hope that the weights and biases approach their optimal values $w \rightarrow \hat w$. 

A conventional neural network (NN) learning a dynamical system might intake a position and velocity $q_i$ and $\dot{q}_i$, output a velocity and acceleration $\dot{q}_o$ and $\ddot{q}_o$, and adjust the weights and biases to minimize the squared difference
\begin{equation}
    C = \left( \dot{q}_i - \dot{q}_o \right)^2 + \left( \ddot{q}_i - \ddot{q}_o \right)^2
\end{equation}
and ensure proper dynamics. After training, NN can intake an initial position and velocity and evolve the system forward in time using a simple Euler update (or some better integration algorithm), as in the top row of Fig.~\ref{xNN}.

\begin{figure}[b!]
	\includegraphics[width=1.0\linewidth]{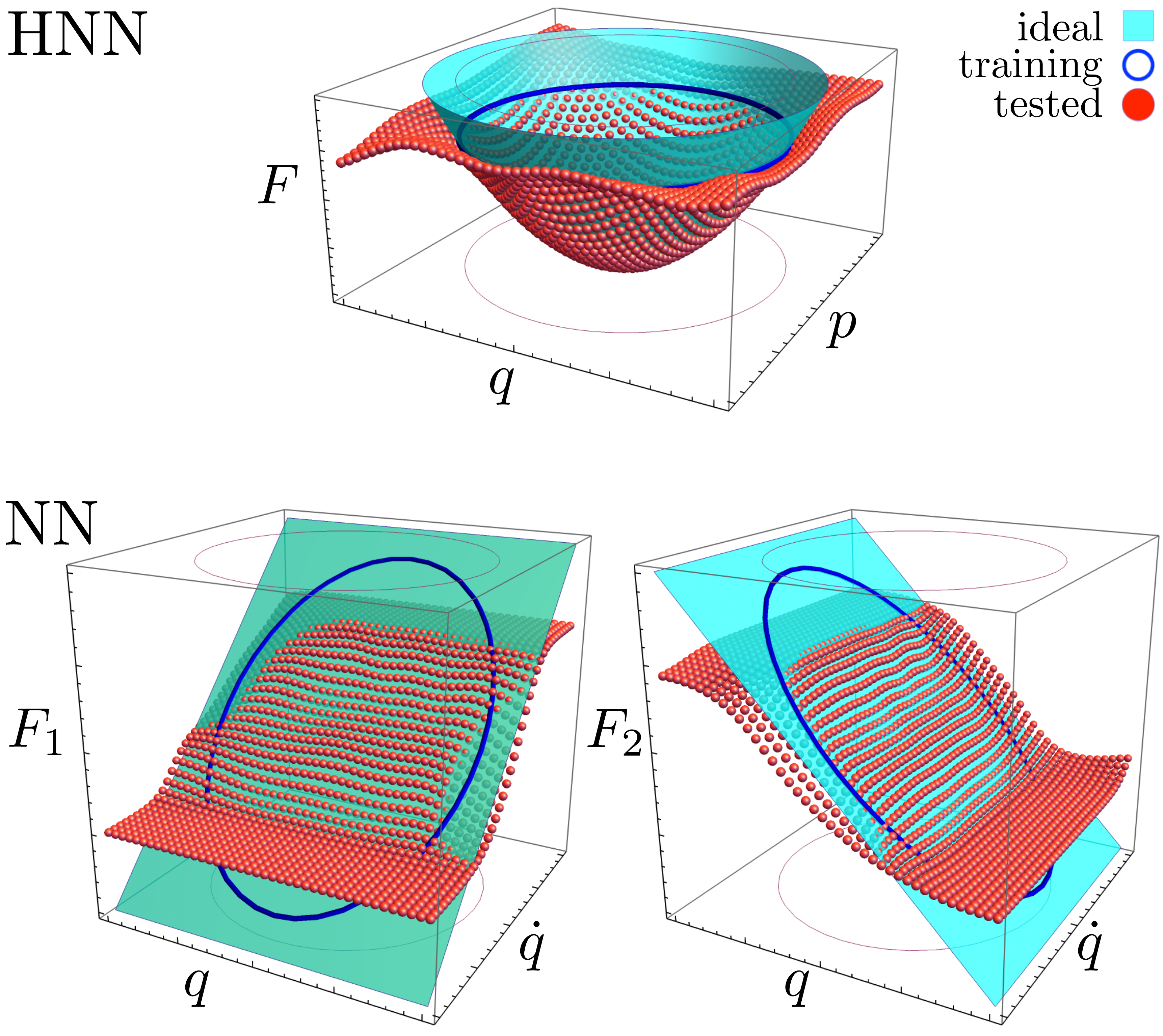} 
	\caption{Mapping compared. HNN maps inputs to the Eq.~\ref{HNNMap} paraboloidal energy surface (top) whose gradient stores the velocity $\dot q$ and force $\dot p = -q$. NN maps linear oscillator inputs to two separate planes (bottom) whose heights are the Eq.~\ref{NNMap} velocity $\dot q$ and acceleration $\ddot q = -q$. Cyan surfaces are targets, training pairs are inside blue circles, red dots are trained tests. Training improves both neural networks.}
	\label{Map}
\end{figure}

\section{Hamiltonian Neural Networks}

To overcome limitations of conventional neural networks, especially when forecasting dynamical systems, recent neural network algorithms have incorporated ideas from physics. In particular, incorporating the symplectic phase space structure of Hamiltonian dynamics has proven very valuable~\cite{HNN,HGN,choudhary}.

A Hamiltonian neural network (HNN) learning a dynamical system intakes position and momentum $q_i$ and $p_i$ but outputs a single energy-like variable $H$, which it differentiates according to Hamilton's recipe
\begin{subequations}
    \begin{align}
        \dot{q}_o &= +\frac{\partial H}{\partial p_i},\\
        \dot{p}_o &= -\frac{\partial H}{\partial q_i},
    \end{align}
\end{subequations}
as in the bottom row of Fig.~\ref{xNN}. Minimizing the HNN cost function
\begin{equation}
    C = \left( \dot{q}_i - \dot{q}_o \right)^2 + \left( \dot{p}_i - \dot{p}_o \right)^2
\end{equation}
then assures symplectic dynamics, including energy conservation and motion on phase space tori. So rather than learning the derivatives, HNN learns the Hamiltonian function which is the {\em generator} of trajectories. Since the same Hamiltonian function generates both ordered and chaotic orbits, learning the Hamiltonian allows the network to forecast orbits outside the training set. In fact it has the capability of forecasting chaos even when trained exclusively on ordered orbit data.

 \begin{figure}[b!]
 	\includegraphics[width=0.85\linewidth]{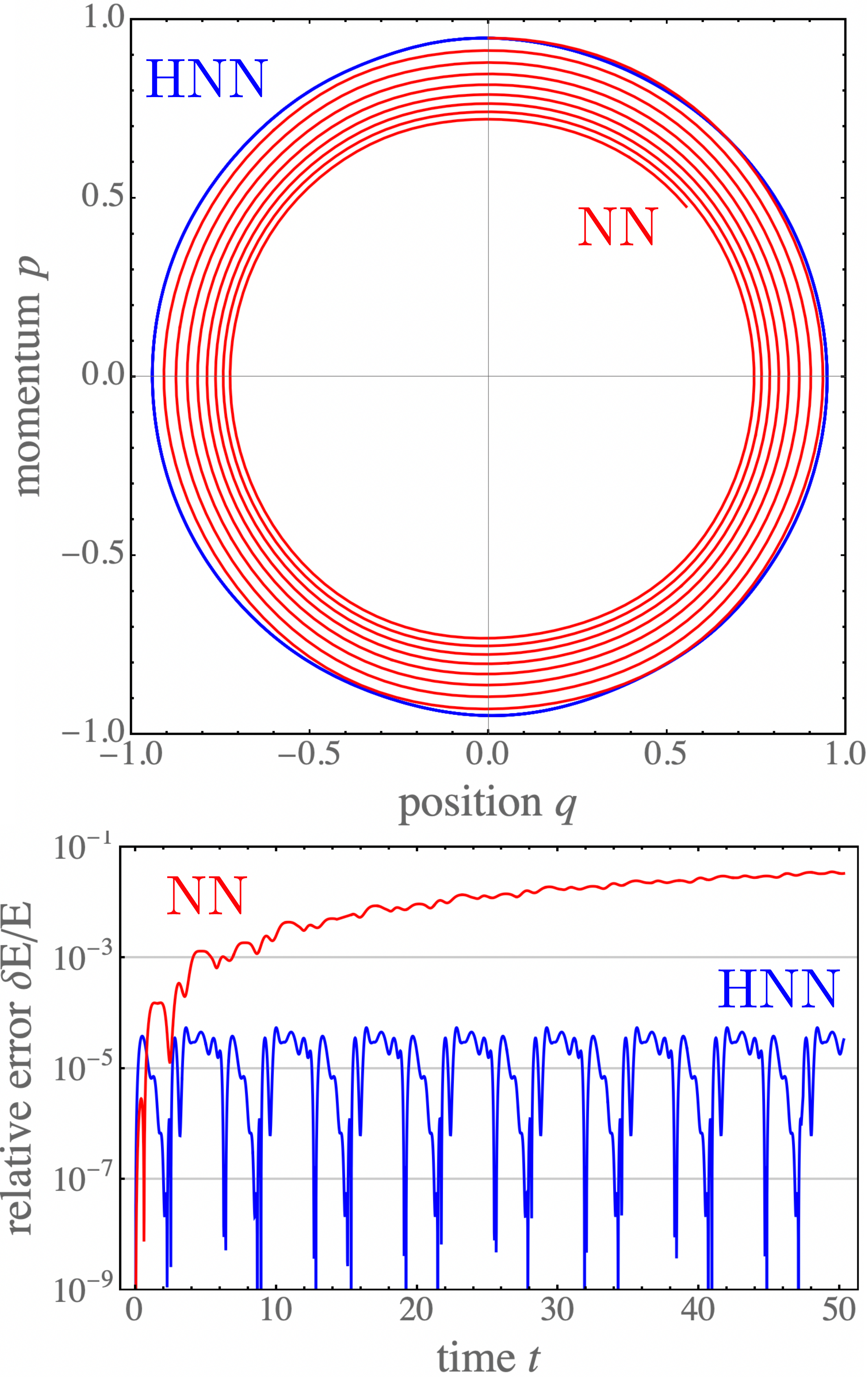}
 	\caption{Linear oscillator forecasting. For the same training parameters and time $0<t<16\pi$, HNN phase space orbit creates a closed circle, while NN phase space orbit slowly spirals in (top). HNN orbit conserves energy much better than NN orbit (bottom).}
 	 \label{Trained}
 \end{figure}
 	
\begin{figure}[tbh]
	\includegraphics[width=0.9\linewidth]{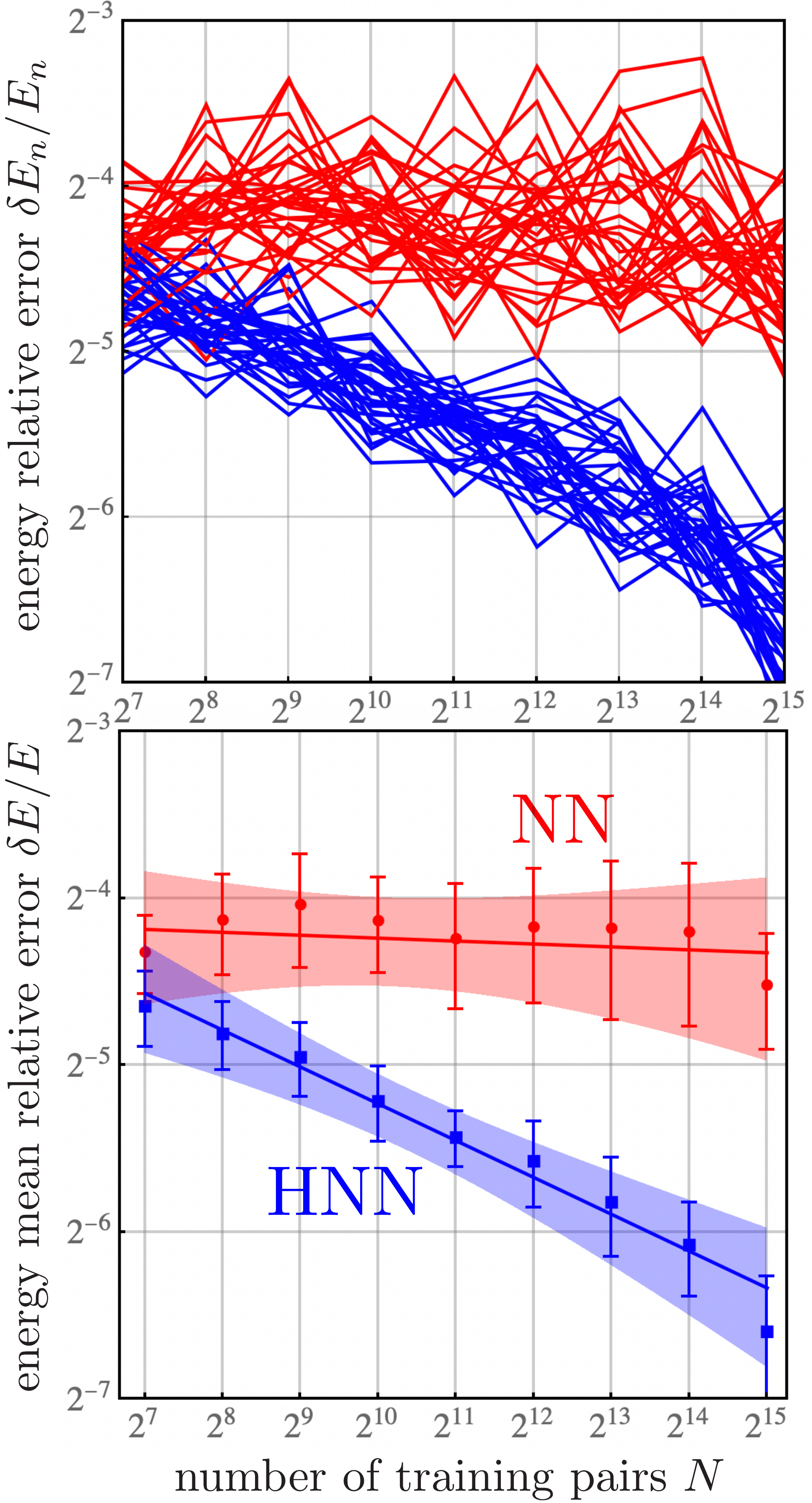} 
	\caption{Learning $d = 6$ linear oscillator. Time-averaged relative error $\delta E / E$ versus number of training pairs $N$ for NN (red) and HNN (blue) for 32 different seeds averaged over 32 different forecasts. Each seed corresponds to a different pseudo-random set of weights and biases $w$. Individual errors (top), mean and standard deviation with 95\% confidence bands (bottom). HNN improves rapidly, like a power law, compared to NN, and has the smaller standard deviation.}
	\label{PowerLaw}
\end{figure}

\begin{figure}[th!]
	\includegraphics[width=0.95\linewidth]{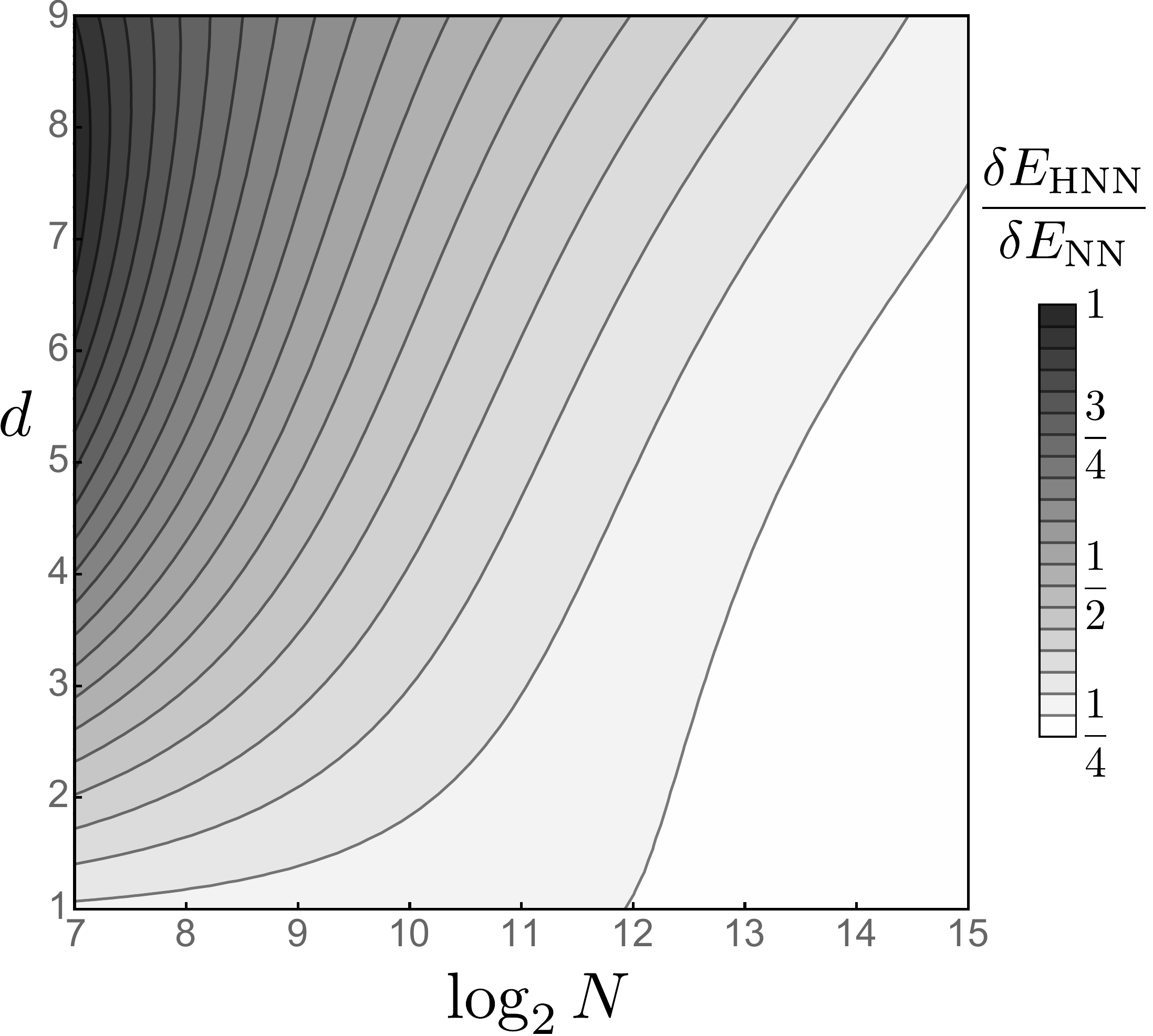} 
	\caption{Smoothed linear forecasting energy error ratio. HNN is up to 4 times better than NN in forecasting the linear oscillator in this domain of number of training pairs $N$ and dimension $d$.}
	\label{LinearErrorRatio}
\end{figure}


\section{Linear Oscillator}

For a simple harmonic oscillator with mass $m = 1$, stiffness $k = 1$,  position $q$, and momentum $p$, the Hamiltonian
\begin{equation}
   H  = \frac{p^2}{2m} + \frac{1}{2} k q^2 = \frac{1}{2}p^2 + \frac{1}{2}q^2,
\end{equation}
so Hamilton's equations
\begin{subequations}
    \begin{align}
        \dot{q} &= +\frac{\partial H}{\partial p} = +p,\\
        \dot{p} &= -\frac{\partial H}{\partial q} = -q
    \end{align}
\end{subequations}
imply the linear equation of motion
\begin{equation}
  \ddot q = -q.
\end{equation}
HHN maps its input to the paraboloid
\begin{equation} \label{HNNMap}
   F_{\hat w} [\{q, p \}] 
   = H 
   =\frac{q^2 + p^2}{2}
   = F,
\end{equation}
but NN maps its input to two intersecting planes
\begin{equation} \label{NNMap}
   F_{\hat w} [\{q, \dot q \}] 
   =\partial_t\{q, \dot q \} 
   =\{\dot q, -q \}
   =\{F_1, F_2 \},
\end{equation}
as illustrated by the Fig.~\ref{Map} cyan surfaces.

We implement the neural networks in Mathematica using symbolic differentiation. HNN and NN train using the same parameters, as in Table~\ref{MMHyperparameterTable}, including 2 hidden layers each of 32 neurons. They train for a range of energies $0<E<1$ and times $0 < t < 2\pi$, and test for times $0 < t < 16\pi$. HNN maps the parabolic energy well, while NN has some problems mapping to the 2 planes, especially for large and small speeds, as in Fig.~\ref{Map}, were cyan surfaces are the ideal targets and red dots are the actual mappings. Training pairs are confined to inside the blue circle, and extrapolation outside is not good in either case, but further training improves both.

Meanwhile, HNN phase space orbit creates a closed circle, while NN phase space orbit slowly spirals in. HNN orbit conserves energy to within $0.01\%$, while NN orbit loses energy by almost $10\%$, for times $0 < t < 16\pi$, as in Fig.~\ref{Trained}.

\begin{figure*}[tbh]
	\includegraphics[width=0.9\linewidth]{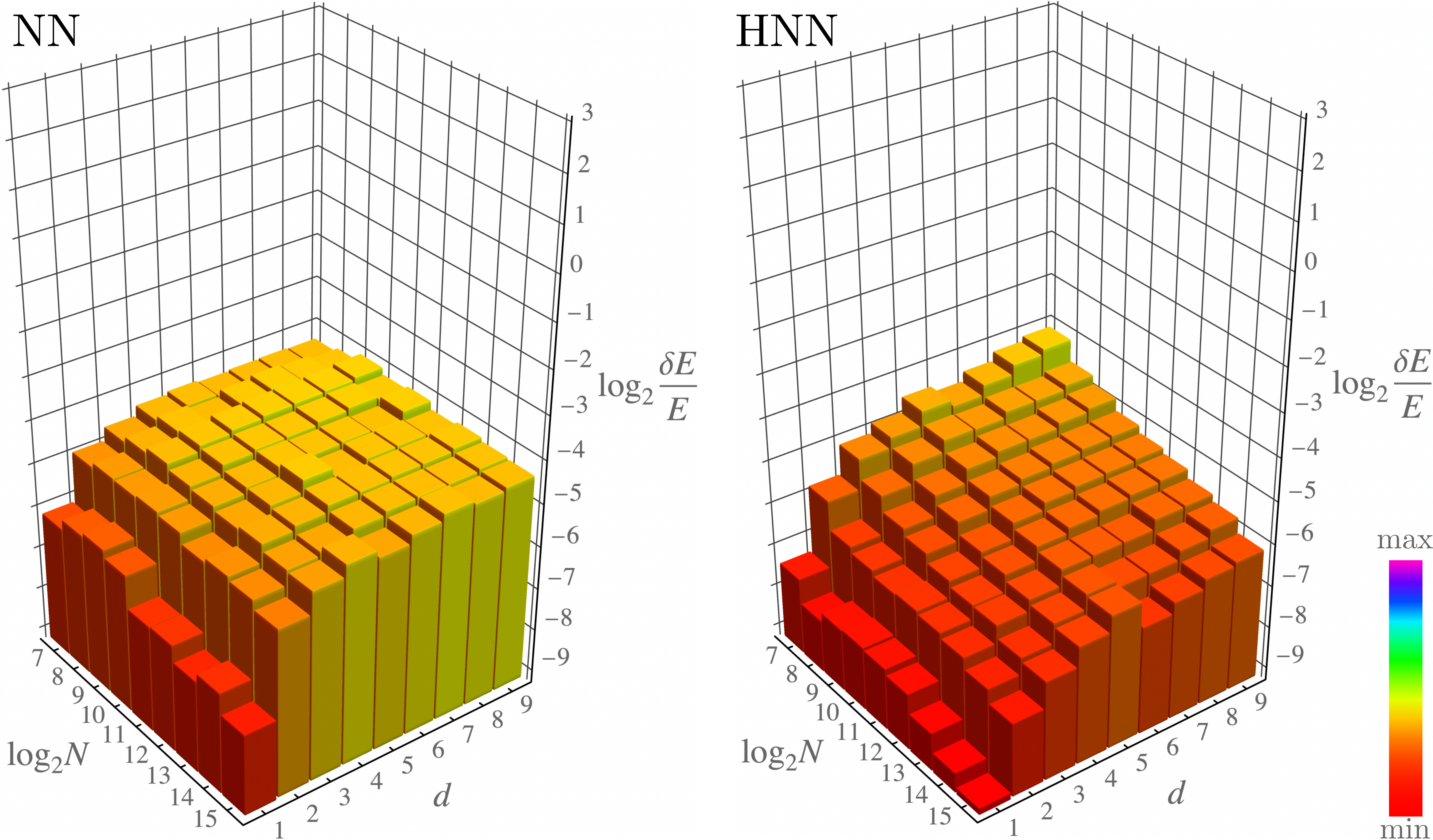} 
	\caption{Linear forecasting. Linear oscillator energy mean relative error  $\delta E/E$ versus number of training pairs $N$ versus dimension $d$, for NN (left) and HNN (right). Each network trains and forecasts 64 times from different initial weights and biases. Rainbow hues code heights. In this domain, HNN forecasts $\lesssim 4$ times better than NN.}
	\label{QuadraticCompare}
\vspace{1.0cm}
	\includegraphics[width=0.9\linewidth]{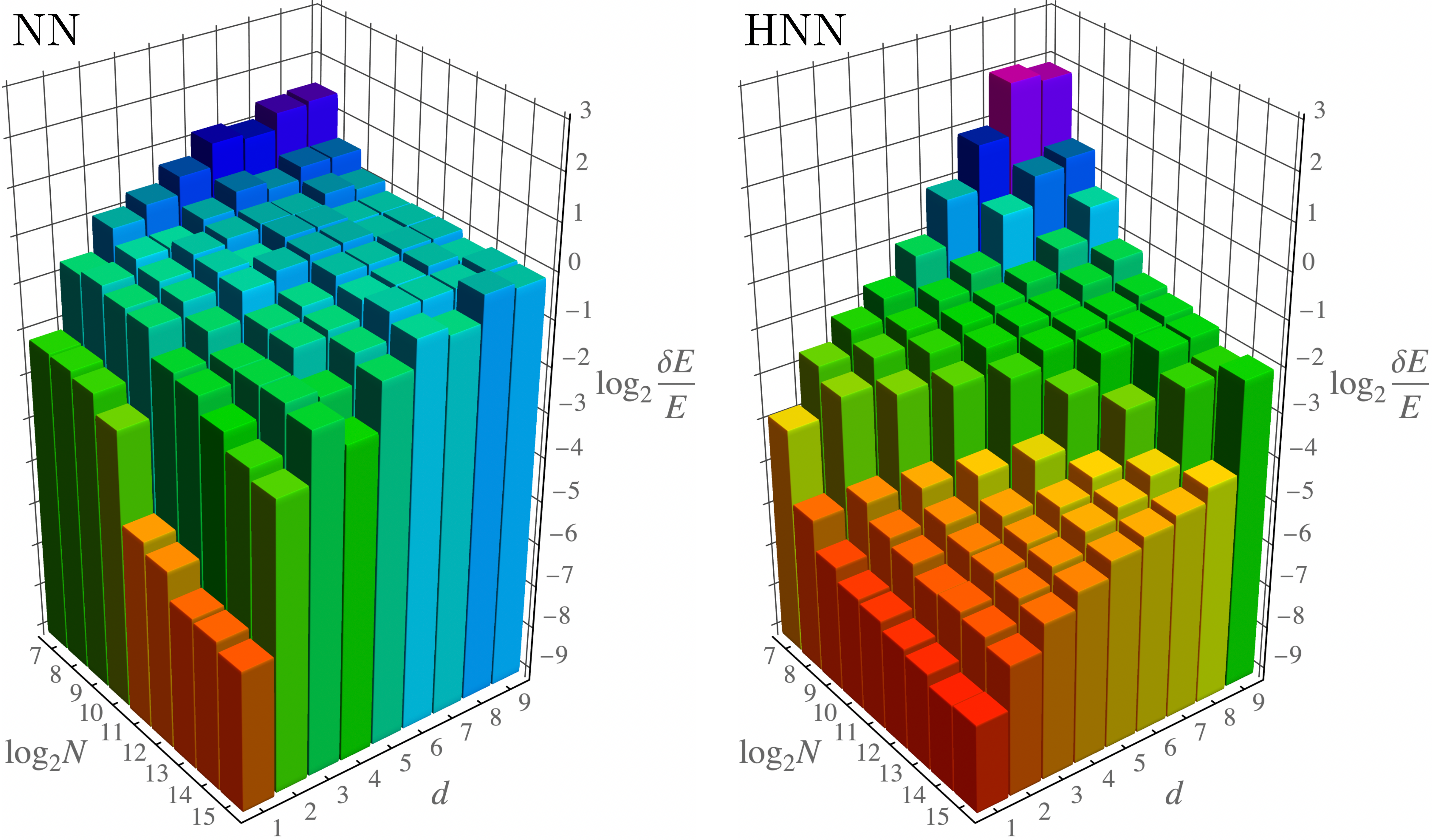} 
	\caption{Nonlinear forecasting. Nonlinear oscillator energy mean relative error  $\delta E/E$ versus number of training pairs $N$ versus dimension $d$, for NN (left) and HNN (right). Each network trains and forecasts 64 times from different initial weights and biases. Rainbow hues code heights with same scale as Fig.~\ref{QuadraticCompare}. In this domain, HNN forecasts $\lesssim 20$ times better than NN.}
	\label{QuarticCompare}
\end{figure*}

\clearpage


\section{Higher Dimensional Oscillators}

More generally, in $d$ spatial dimensions and $2d$ phase space, the quadratic oscillator Hamiltonian
\begin{equation}
      H_2 = \sum_{n=1}^d\left(\frac{1}{2}p_n^2 + \frac{1}{2}q_n^2 \right)
\end{equation}
has a linear restoring force, but the $d$-dimensional quartic oscillator
\begin{equation}
   H_4 = \sum_{n=1}^d\left(\frac{1}{2}p_n^2 + \frac{1}{4}q_n^4 \right)
\end{equation}
has a nonlinear restoring force.

We implement the neural networks in Python using automatic differentiation. HNN and NN train with the same parameters, some of which are optimized as in Table~\ref{PythonHyperparameterTable}, including 2 hidden layers of 32 neurons. They train for a range of energies $0 < E < 1$ and times $0 < t < 100$. We compute the energy mean relative error $\delta E/E$ of each forecasted orbit, which we further average over 64 training sessions, each starting with a unique set of initial weights and biases $w$. For each dimension $d$, we plot the error $\delta E/E$  versus the number of training pairs $N$.

Figure~\ref{PowerLaw} summarizes the linear oscillator results for dimension $d = 6$.  Raw errors (top) suggest variance, and mean errors with 95\% confidence band (bottom) indicate variance. HNN has smaller variance and improves dramatically with increasing training, in this case like the power law
\begin{equation}
  \frac{\delta E}{E} \sim 0.12 N^{-0.22}.
\end{equation}

We repeat the forecasting error analysis for dimensions $1 \le d \le 9$. HNN maintains its forecasting edge over NN in higher dimensions, as summarized by the smoothed Fig.~\ref{LinearErrorRatio} contour plot. Each network trains 32 times from different initial weights and biases and then forecasts 32 different orbits. Figure~\ref{QuadraticCompare} heights and rainbow hues code energy mean relative errors. NN rapidly loses accuracy with dimension for all tested training pairs. HHN slowly loses accuracy with dimension but recovers it with training pairs.

Next we repeat the forecasting error analysis for nonlinear oscillators, as in Fig.~\ref{QuarticCompare}. Although nonlinear oscillator is harder to learn,  HNN still delivers good forecasts for sufficiently many training pairs.


\section{Bistable Chain}
\begin{figure}[t!]
	\includegraphics[width=1.0\linewidth]{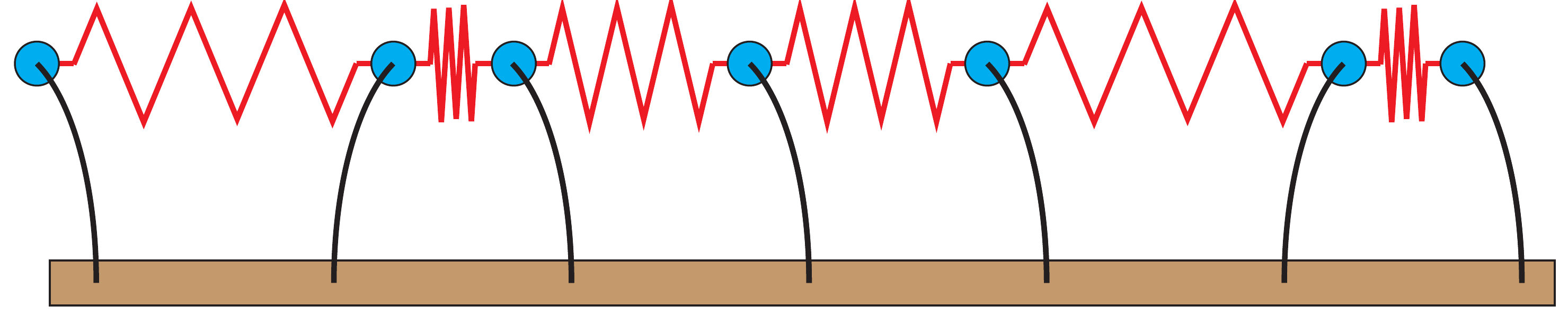} 
	\caption{Bistable chain.  Hacksaw blades (black) stuck vertically into a piece of wood (brown) with small masses (blue) attached at the tops. Vertical is an unstable equilibrium and each blade sags left or right forming a bistable system. (The width of the blade prevents it from sagging into or out of the drawing.) Linear springs (red) couple the masses. }
	\label{DuffingArrayModel}
\end{figure}

Finally, consider a chain of coupled bistable oscillators, as in Fig.~\ref{DuffingArrayModel}, where top-heavy hacksaw blades joined by Hooke's law springs swing back-and-forth between their dual sagging equilibria. Model each blade by the nonlinear spring force
\begin{equation}
  f [q]= a q - b q^3
\end{equation}
with $a,b >0$. The corresponding potential
\begin{equation}
 V[q] = -\int f[q]\, dq = -\frac{1}{2}a q^2 + \frac{1}{4}b q^4
\end{equation}
has an unstable equilibrium at $q = 0$ and stable equilibria at $q = \pm \sqrt{a/b}$. Couple adjacent masses by linear springs of stiffness $\kappa$. For $d$ identical masses $m=1$, the Hamiltonian
\begin{equation}
H_c = \sum_{n=1}^d \left( \frac{1}{2}p_n^2+V[q_n]+\frac{1}{2} \kappa \left(q_n-q_{n+1}\right)^2 \right).
\end{equation}
Hamilton's equations imply
\begin{subequations}
 \begin{align}
\dot{q}_n &= p_n,\\
\dot{p}_n &= V'[q_n]+ \kappa \left( q_{n-1} - q_n \right) + \kappa \left( q_{n+1} - q_{n} \right)\nonumber\\
&= a q_n - b q_n^3 + \kappa\left( q_{n-1} - 2q_n  + q_{n+1}\right).
\end{align}\end{subequations}
Enforce free boundary conditions by demanding
\begin{subequations}
\begin{align}
q_{0}=q_1,\\
q_{d+1}=q_d.
\end{align}\end{subequations}

\begin{figure*}[hbt!]
	\includegraphics[width=0.9\linewidth]{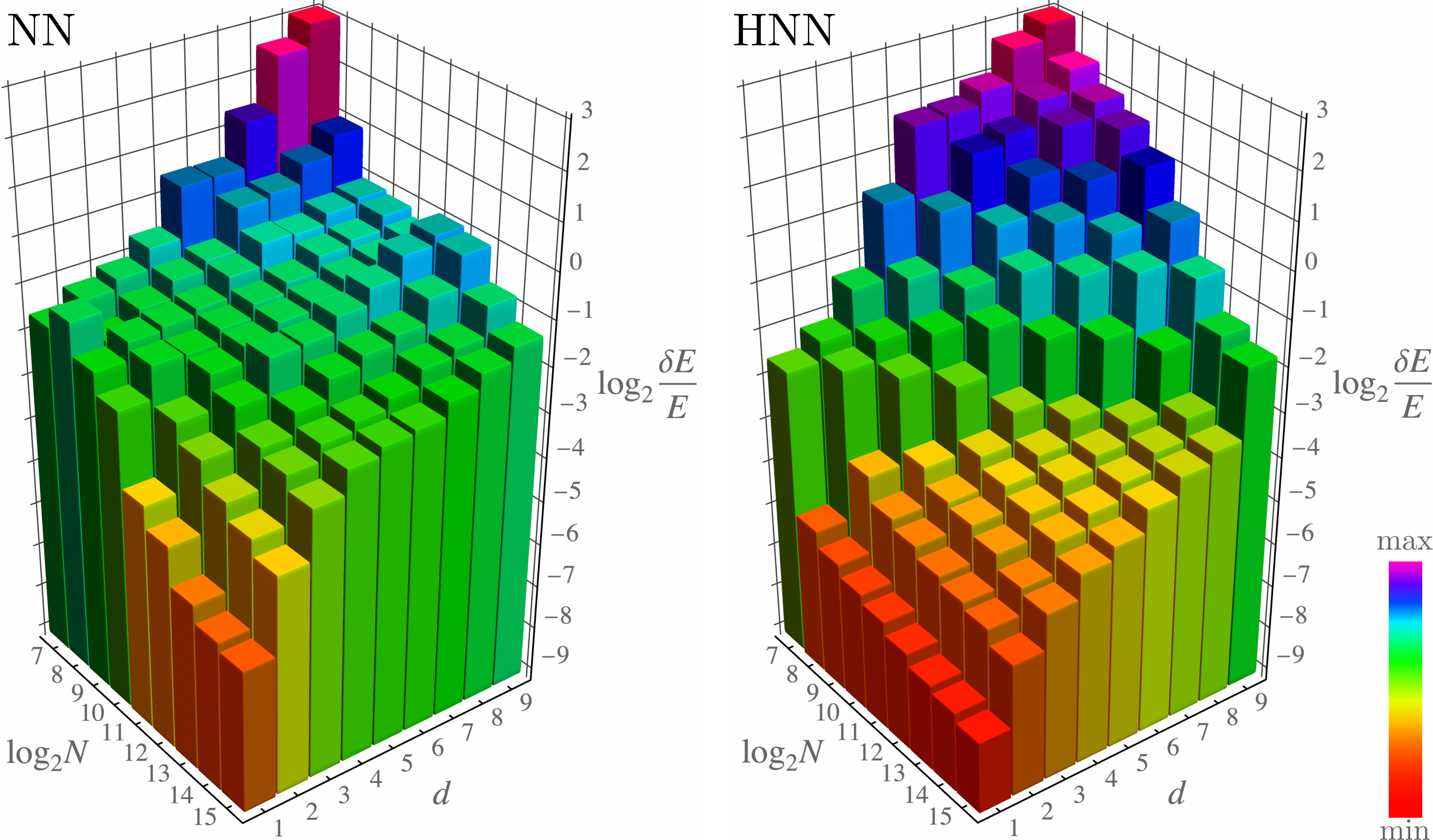} 
\caption{Bistable chain forecasting. Bistable chain energy mean relative error $\delta E/E$ versus number of training pairs $N$ versus dimension $d$, for NN (left) and HNN (right). Each network trains and forecasts 64 times from different initial weights and biases. Heights and rainbow hues code energy errors with same scale as Figs.~\ref{QuadraticCompare}-\ref{QuarticCompare}. In this domain, HNN forecasts $\lesssim 9$ times better than NN.}
	\label{BistableChain}
\end{figure*}

As with the uncoupled higher-dimensional systems, for sufficiently many training pairs, HNN significantly outperforms NN in forecasting the bistable chain, as in Fig.~\ref{BistableChain}. For sufficiently \textit{few} training pairs, NN appears to occasionally outperform HNN. While HNN must learn to map just the single energy surface, it must learn the surface well enough to estimate its gradient (which stores the velocities and forces), and this requires sufficient training. But when NN outperforms HNN, as in the Fig.~\ref{BistableChain} low-$N$ high-$d$ back corner, \textit{neither} network learns well, and the best strategy is to increase the number of training pairs and use HNN.


\section{Conclusions}

Artificial neural networks can forecast dynamical systems by continuously adjusting their weights and biases. This analog skill is complementary to the algebraic skill of systems like Eureqa~\cite{Eureqa1,Eureqa2} and SINDy~\cite{SINDy1,SINDy2} that learn symbolic equations of motion.

We have examined the ability of recently introduced Hamiltonian neural networks~\cite{HNN,HGN,mattheakis2019,Bertalan2019,Bondesan2019,choudhary} to learn high-dimensional dynamics and gauged its effectiveness compared to conventional neural networks. The test-bed of our investigations were linear and nonlinear oscillators and a chain of coupled bistable oscillators. Of the three error metrics, neural network cost function $C$, forecasted energy error $\delta E / E$, and forecasted state error $\delta r$, where the state $\vec r = \{q_n,p_n\}$, we expect small $\delta r$ implies small $\delta E/E$ implies small $C$, but not conversely. We chose the forecasted energy error to quantitatively assess the potency of the algorithms, as this metric reflects the importance of energy conservation, is fast to compute, and is a good indicator of forecasting power. 

Further we introduced the idea of construing neural networks as nonlinear mappings, and this provided insights into their capabilities. In particular it helped elucidate the underlying reason why Hamiltonian neural networks can  outperform conventional neural networks when forecasting dynamical systems. For instance, the linear oscillator offered an excellent example, because it involved mappings that are simple geometrical surfaces, thus illustrating clearly how the advantage of HNN in higher dimensions is accentuated.

What if the number of training pairs cannot be increased due to a paucity of data, for example? Future work will include systematically increasing the depth and breadth of the neural network (as well as other hyperparameters) to try to improve the forecasting as the dimension of the dynamics increases. How does the forecasting vary as the chaos of the system varies, as measured both by largest Lyapunov exponent and number of positive exponents (chaos versus hyperchoas)? We hope to systematically vary the quality and quantity of chaos and record the effects on training and forecasting using all three error metrics: cost, energy, and state.

The basic principle underlying the success of HNN is the fact that a {\em single} function, the Hamiltonian, is a generator of the entire phase space dynamics, in any number of dimensions. So the task of learning is confined to learning this single powerful function, irrespective of dimensionality, as the evolution of {\em all} variables are determined through the derivatives of this single function. We demonstrated that by simply incorporating this broad physics principle, one gains significant power in forecasting complex dynamical systems. Specifically, the relative error decreases as a power-law of the number of training pairs for HNN even for higher dimensional systems. In contrast, conventional NN requires significantly more training pairs to learn the same dynamics, especially in higher dimensions. A neural network enhanced by a basic formalism from physics, the Hamiltonian formalism, is better equipped to handle real-world mechanical systems, which are necessarily multi-component and thus high-dimensional. 

\section*{Acknowledgments}

This research was supported by ONR grant N00014-16-1-3066 and a gift from United Therapeutics. J.F.L. thanks The College of Wooster for making possible his sabbatical at NCSU. S.S. acknowledges support from the J.C. Bose National Fellowship (SB/S2/JCB-013/2015).

\appendix*

\section{Neural Network Training Details}
In feed-forward artificial neural networks, the activity of neurons in one layer
\begin{equation}
    a_\ell \overset{\text{vec}}{=}  \sigma \left[ w_\ell\, a_{\ell-1}+ b_\ell \right]
\end{equation}
is a sigmoid function of a linear combination of the activities in the previous layer. The concatenation of such functions eliminates the activities and produces the nonlinear input-output Eq.~\ref{nonlinearEq}, where the weights and biases $w = \{w_\ell,b_\ell\}$.

For robustness, we implemented our neural networks in two different environments; Mathematica using symbolic differentiation, for Fig.~\ref{Map} and Fig.~\ref{Trained}, and Python using automatic differentiation, for the other figures. Table~\ref{MMHyperparameterTable} and Table~\ref{PythonHyperparameterTable} summarize our training parameters. For Python, the Adam optimizer algorithm~\cite{Adam2014} modified the base learning rate according to the gradient of the loss function with respect to the weights and biases. All simulations ran on desktop computers.

\begin{table}[htb!] 
	\caption{Mathematica neural network training parameters.}
	\label{MMHyperparameterTable}
	\begin{ruledtabular}
		\begin{tabular}{lll} 
			symbol & name & value \\
			\hline
			$N_\ell$ & hidden layers (depth) & $2$\\
			$N_1$ & neurons per layer (width) & $2^5$\\
			$N_n = N_\ell N_1$ & neurons & $2^6$ \\
			\hline
			$T$ & training time per orbit & $2\pi$\\
			$\Delta t$ & sampling time & $2\pi / 2^7$\\
			$T/\Delta t$ & samples per orbit & $2^7$\\
			$N_o$ & orbits & $2^7$\\
			$N_t = N_o T/\Delta t$ & training pairs & $2^{14}$\\
			\hline
			$N_b$ & batches & $2^{14}$ \\
			$N_s$ & batch size = pairs per batch & $1$ \\
			$N_t = N_b N_s$ & training pairs & $2^{14}$ \\
			\hline
			$N_e$ & epochs = data visits & $2^4$ \\
			$N_i = N_e N_t$ & total inputs & $2^{18}$ \\
			\hline
			$\Delta E$ & training energy range & $[10^{-3},1]$\\
			$\eta$ & learning rate & $10^{-3}$ \\  
			$\sigma$ & activation function & $\tanh$ \\  
		\end{tabular}
	\end{ruledtabular}
\vspace{0.0cm}
	\caption{Python neural network training parameters.}
	\label{PythonHyperparameterTable}
	\begin{ruledtabular}
		\begin{tabular}{lll} 
			symbol & name & value \\
			\hline
			$N_\ell$ & hidden layers (depth) & $2$\\
			$N_1$ & neurons per layer (width) & $2^5$\\
			$N_n = N_\ell N_1$ & neurons & $2^6$ \\
			\hline
			$T$ & training time per orbit & $10^2$\\
			$\Delta t$ & sampling time & $0.1$\\
			$T/\Delta t$ & samples per orbit & $10^3$\\
			\hline
			$N_b$ & batches & $2^{15}$ \\
			$N_s$ & batch size = pairs per batch & $1$ \\
			$N_t = N_b N_s$ & training pairs & $\le 2^{15}$ \\
			\hline
			$N_e$ & epochs = data visits & $2^4$ \\
			$N_i = N_e N_t$ & total inputs & $\le 2^{19}$ \\
			\hline
			$\Delta E$ & training energy range & $[0,100]$\\
			$\eta$ & learning rate & $10^{-3}$ \\  
			$\sigma$ & activation function & $\tanh$ \\  
		\end{tabular}
	\end{ruledtabular}
\end{table}

\clearpage

\bibliography{LearningChaos}

\end{document}